\documentclass[11pt,letterpaper]{article}
\usepackage{emnlp2016}
\usepackage{times}
\usepackage{latexsym}
\usepackage{booktabs}
\usepackage{graphics}
\usepackage{array}
\usepackage{tabularx}
\usepackage{graphicx}
\usepackage{multirow}
\usepackage{amsmath}
\usepackage[noend]{myalgorithmic}
\usepackage{myalgorithm}
\usepackage{amssymb}
\usepackage{nicefrac}
\usepackage{hyperref}

\emnlpfinalcopy

\def\emnlppaperid{***}

\title{De-Conflated Semantic Representations}

\author{Mohammad Taher Pilehvar \and Nigel Collier\\
 Language Technology Lab\\
 University of Cambridge\\
 Cambridge, UK\\
 {\tt \{mp792,nhc30\}@cam.ac.uk}}

\date{}

\begin{document}

\maketitle

\begin{abstract}

One major deficiency of most semantic representation techniques is that they usually model a word type as a single point in the semantic space, hence conflating all the meanings that the word can have.
Addressing this issue by learning distinct representations for individual meanings of words has been the subject of several research studies in the past few years.
However, the generated sense representations are either not linked to any sense inventory or are unreliable for infrequent word senses.
We propose a technique that tackles these problems by de-conflating the representations of words based on the deep knowledge it derives from a semantic network.
Our approach provides multiple advantages in comparison to the past work, including its high coverage and the ability to generate accurate representations even for infrequent word senses.
We carry out evaluations on six datasets across two semantic similarity tasks and report state-of-the-art results on most of them.

\end{abstract}

\section{Introduction}


Modeling the meanings of linguistic items in a machine-interpretable form, i.e., semantic representation, is one of the oldest, yet most active, areas of research in Natural Language Processing (NLP).
The field has recently experienced a resurgence of research interest with the new blood injected in its veins by neural network-based models that view the representation task as a language modeling problem and learn dense representations (usually referred to as embeddings) by efficiently processing massive amounts of texts. However, either in its conventional count-based form \cite{TurneyPantel:2010} or the recent predictive approach, the prevailing objective of representing each word type as a single point in the semantic space has a major limitation: it ignores the fact that words can have multiple meanings and conflates all these meanings into a single representation. 
This objective can have negative impacts on accurate semantic modeling, e.g., semantically unrelated words that are synonymous to different senses of a word are pulled towards each other in the semantic space \cite{Neelakantanetal:2014}. 


Recently, there has been a growing interest in addressing the meaning conflation deficiency of word representations. 
A series of techniques tend to associate a word to multiple points in the semantic space by clustering its contexts in a given text corpus and learning distinct representations for individual clusters \cite{ReisingerMooney:2010,Huangetal:2012}.
Though, these techniques usually assume a fixed number of word senses per word type, disregarding the fact that the number of senses of a word can range from one (monosemy) to dozens.
\newcite{Neelakantanetal:2014} tackled this issue by allowing the number to be dynamically adjusted for each word during training.
However, the approach and all the other clustering-based techniques still suffer from the fact that the computed sense representations are not linked to any sense inventory, a linking which would require large amounts of sense-annotated data \cite{Agirreetal:2006}.
In addition, because of their dependence on knowledge derived from a text corpus, these techniques are generally unable to learn accurate representations for word senses that are infrequent in the underlying corpus.

Knowledge-based techniques tackle these issues by deriving sense-specific knowledge from external sense inventories, such as WordNet \cite{Fellbaum:98}, and  learning representations that are linked to the sense inventory.
These approaches either use sense definitions and employ Word Sense Disambiguation (WSD) to gather sense-specific contexts \cite{Chenetal:2014,Iacobaccietal:2015} or take advantage of the properties of WordNet, such as synonymy and direct semantic relations \cite{RotheSchutze:2015}.
However, the non-optimal WSD techniques and the shallow utilization of knowledge from WordNet do not allow these techniques to learn accurate and high-coverage semantic representations for all senses in the inventory.

We propose a technique that de-conflates a given word representation into its constituent sense representations by exploiting deep knowledge from the semantic network of WordNet.
Our approach provides the following three main advantages in comparison to the past work:
(1) our representations are linked to the WordNet sense inventory and, accordingly, the number of senses for a word is a dynamic parameter which matches that defined by WordNet;
(2) the deep exploitation of WordNet's semantic network allows us to obtain accurate semantic representations, even for word senses that are infrequent in generic text corpora; and
(3) our methodology involves only minimal parameter tuning and can be effectively applied to any sense inventory that is viewable as a semantic network and to any word representation technique.
We evaluate our sense representations in two tasks: word similarity (both in-context and in-isolation) and cross-level semantic similarity.
Experimental results show that the proposed technique can provide consistently high performance across six datasets, outperforming the recent state of the art on most of them.


\section{De-Conflated Representations}

\paragraph{Preliminaries.}

Our proposed approach takes a set of pre-trained word representations and uses the graph structure of a semantic lexical resource in order to de-conflate the representations into those of word senses.
Therefore, our approach firstly requires a set of pre-trained word representations (e.g., word embeddings). 
Any model that maps a given word to a fixed-size vector representation (i.e., vector space model) can be used by our approach. 
In our experiments, we opted for a set of publicly available word embeddings (cf. \S\ref{sec:exp_setup}).

Secondly, we require a lexical resource whose semantic relations allow us to view it as a graph $G=(V,E)$ where each vertex in the set of vertices $V$ corresponds to a concept and edges in $E$ denote lexico-semantic relationships among these vertices.
Each concept $c \in V$ is mapped to a set of word senses by a mapping function $\mu(c) : c \rightarrow \{s_1,\dots,s_l\}$.
WordNet, the \textit{de facto} community standard sense inventory, is a suitable resource that satisfies these properties.
WordNet can be readily represented as a semantic graph in which vertices are synsets and edges are the semantic relations that connect these synsets (e.g., hypernymy and meronymy).
The mapping function in WordNet maps each synset to the set of synonymous words it contains (i.e., word senses).

\subsection{Overview of the approach}

Our goal is to compute a semantic representation that places a given word sense in an existing semantic space of words.
We achieve this by leveraging word representations as well as the knowledge derived from WordNet.
The gist of our approach lies in its computation of a list of \textit{sense biasing words} for a given word sense.
To this end, we first analyze the semantic network of WordNet and extract a list of most representative words that can effectively pinpoint the semantics of individual synsets (\S\ref{sec:filtering_words}).
We then leverage an effective technique which learns semantic representations for individual word senses by placing the senses in the proximity of their corresponding sense biasing words (\S\ref{sec:learn_rep}).

\subsection{Determining sense biasing words}
\label{sec:filtering_words}

Algorithm \ref{alg1} shows the procedure we use to extract from WordNet a list of sense biasing words for a given target synset $y_t$.
The algorithm receives as its inputs the semantic graph of WordNet and the mapping function $\mu(\cdot)$, and outputs an ordered list of biasing words $\mathcal{B}_t$ for $y_t$.
The list comprises the most semantically-related words to synset $y_t$ which can best represent and pinpoint its meaning.
We leverage a graph-based algorithm for the computation of the sense biasing words. 

\begin{algorithm}[t!]
\caption{Get sense biasing words for synset $y_t$} 
\label{alg1}                         
\small
\begin{algorithmic}[1]
  \REQUIRE Graph $G=(V,E)$ of vertices $V=\{y_i\}^m_{i=1}$ (of $m$ synsets) and edges $E$ (semantic relationships between synsets)
  \REQUIRE Function $\mu(y_i)$ that returns for a given synset $y_i$ the words it contains
  \REQUIRE Target synset $y_t \in V$ for which a sense biasing word sequence is required
  
  \ENSURE The sequence $\mathcal{B}_t$ of sense biasing words for synset $y_t$


    \STATE $\mathcal{B}_t$ $\leftarrow$ $()$
    \FORALL {word $w$ in $\mu(y_t)$}   
            \STATE {$\mathcal{B}_t$ $\leftarrow$ $\mathcal{B}_t \cup (w)$} \label{line:init}
    \ENDFOR
    \FOR {$y_i \in V$ : $y_i \ne y_t$}  
        \STATE {$p_i$ $\leftarrow$ \textsc{PersonalizedPageRank}$(y_i, y_t, G)$} \label{line:ppr}
    \ENDFOR
    
    \STATE {$(y^*_h)^{m-1}_{h=1}$} $\leftarrow$ \textsc{Sort}($V$ \textbackslash $\{y_t\}$) according to scores $p_i$ \label{line:sort}
    \FOR {$h$ : 1 to $m-1$}       \label{line:beg}
        \FORALL {word $w$ in $\mu(y^*_h)$}
            \IF {$w \notin \mathcal{B}_t$}
            \STATE {$\mathcal{B}_t$ $\leftarrow$ $\mathcal{B}_t \cup (w)$} 
            \ENDIF
        \ENDFOR
    \ENDFOR     \label{line:end}
  
    \RETURN sequence $\mathcal{B}_t$ \label{line:return}

\end{algorithmic}
\end{algorithm}

Specifically, we use the Personalized PageRank \cite[PPR]{Haveliwala:2002} algorithm which has been extensively used by several NLP applications \cite{Yehetal:2009,NiemannGurevych:2011,Agirreetal:2014}. 
To this end, we first represent the semantic network of WordNet as a row-stochastic transition matrix $\mathbf{M} \in \mathbb{R}^{m \times m}$ where $m$ is the number of synsets in WordNet ($|V|$).
The cell $M_{ij}$ of $\mathbf{M}$ is set to the inverse of the degree of $i$ if there is a semantic relationship between synsets $i$ and $j$ and to zero otherwise.
We compute the PPR distribution for a target synset $y_t$ by using the power iteration method  $\mathcal{P}^{t+1} = (1-\sigma)\mathcal{P}^0 + \sigma \mathbf{M} \mathcal{P}^t$, where $\sigma$ is the damping factor (usually set to 0.85) and $\mathcal{P}^0$ is a one-hot initialization vector with the corresponding dimension of $y_t$ being set to 1.0.
The weight $p_i$ in line \ref{line:ppr} is the value of the $i^{th}$ dimension of the PPR vector $\mathcal{P}$ computed for the synset $y_t$.
This weight can be seen as the importance of the corresponding synset of the $i^{th}$ dimension (i.e., $y_i$) to $y_t$.
When applied to a semantic network, such as the WordNet graph, this importance can be interpreted as semantic relevance.
Hence, the value of $p_i$ denotes the extent of semantic relatedness between $y_i$ and $y_t$.
We use this notion and retrieve a list of most semantically-related words to $y_t$.
To achieve this, we sort the synsets $\{y^* \in V: y^* \ne y_t\}$ according to their PPR values $\{p_i\}_{i=1}^{m-1}$ (line \ref{line:sort}).
We then iterate (lines \ref{line:beg}-\ref{line:end}) the sorted list $(y^*)$ and for each synset $y^*_h$ append the list $\mathcal{B}_t$ with all the words in $y^*_h$ (i.e., $\mu(y^*_h)$).
However, in order to ensure that the words in the target synset $y_t$ appear as the most representative words in $\mathcal{B}_t$, we first assign
these words to the list (line \ref{line:init}).
Finally, the algorithm returns the ordered list $\mathcal{B}_t$ of sense biasing words for the target synset $y_t$. 

Table \ref{table:alg1_output} shows a sample of top biasing words extracted for the two senses of the noun \textit{digit}: the numerical and the anatomical senses.\footnote{The first and third senses of the noun \textit{digit} in WordNet 3.0.}
We explain in \S\ref{sec:learn_rep} how we use the sense biasing lists to learn sense-specific representations.
Note that the size of the list is equal to the total number of strings in WordNet.
However, we observed that taking a very small portion of the top-ranking elements in the lists is enough to generate representations that perform very similarly to those generated when using the full-sized lists (please see \S\ref{sec:down_sizing}).

\begin{table}
\begin{center}
\small
\scalebox{0.85}
{
\begin{tabular}{ll}
\toprule

\bf \#  &\bf   Sense biasing words     \\
\midrule
1   &   dactyl, finger, toe, thumb, pollex, body\_part, nail, minimus,\\
    &   tarsier, webbed, extremity, appendage \\
\midrule
2   &   figure, cardinal\_number, cardinal, integer, whole\_number, \\
    &   numeration\_system, number\_system, system\_of\_numeration, \\
    &   large\_integer, constituent, element, digital  \\

\bottomrule

\end{tabular}
}
\end{center}
\caption{\label{table:alg1_output} The top sense biasing words for the synsets containing the anatomical (\#1) and numerical (\#2) senses of the noun \textit{digit}.}
\end{table}

\subsection{Learning sense representations}
\label{sec:learn_rep}

Let $\mathcal{V}$ be the set of pre-trained $d$-dimensional word representations. 
Our objective here is to compute a set $\mathcal{V}^* = \{v^*_{s_1},\dots, v^*_{s_n}\}$ of representations for $n$ word senses $\{s_1,\dots, s_n\}$ in the same $d$-dimensional semantic space of words.
We achieve this for each sense $s_i$ by de-conflating the representation $v_{s_i}$ of its corresponding lemma and biasing it towards the representations of the words in $\mathcal{B}_i$.
Specifically, we obtain a representation  $v^*_{s_i}$ for a word sense $s_i$ by solving:
\begin{equation}
\label{main_formula}
\arg\min_{v^*_{s_i}} \alpha \; d(v^*_{s_i},v_{s_i}) + \\ \sum_{b_{ij} \in \mathcal{B}_i}{\delta_{ij} \; d(v^*_{s_i},v_{b_{ij}})}
\end{equation}
%
\noindent where $v_{s_i}$ and $v_{b_{ij}}$ are the respective word representations ($\in \mathcal{V}$) of the lemma of $s_i$ and the $j^{th}$ biasing word in the list of biasing words for $s_i$, i.e, $\mathcal{B}_i$. 
The distance $d(v,v')$ between vectors $v$ and $v'$ is measured by squared Euclidean distance $\lVert v - v' \lVert^2 = \sum_{k} (v_k - v'_k)^2$.
The first term in Formula \ref{main_formula} requires the representation of the word sense $s_i$ (i.e., $v^*_{s_i}$) to be similar to that of its corresponding lemma, i.e., $v_{s_i}$, whereas the second term encourages $v^*_{s_i}$ to be in the proximity of its biasing words in the semantic space.
The above criterion is similar to the frameworks of \newcite{DasSmith:2011} and \newcite{Faruquietal:2015} which, though being convex, is usually solved for efficiency reasons by an iterative method proposed by \newcite{Bengioetal:2007}.
Following these works, we obtain the below equation for computing the representation of a word sense $s_i$:
\begin{equation}
    v^*_{s_i} = \frac{\alpha v_{s_i} + \sum_{b_{ij} \in \mathcal{B}_i}{\delta_{ij} v_{b_{ij}}}}{\alpha + \sum_{j}{\delta_{ij}}}.
\end{equation}


We define $\delta_{ij}$ as  $\frac{e^{-\lambda r(i,j)}}{|\mathcal{B}_i|}$ where $r(i,j)$ denotes the rank of the word $b_{ij}$ in the list $\mathcal{B}_i$.
This is essentially an exponential decay function that gives more importance to the top-ranking biasing words for $s_i$.
The hyperparameter $\alpha$ denotes the extent to which $v^*_{s_i}$ is kept close to its corresponding lemma representation $v_{s_i}$.
Following \newcite{Faruquietal:2015}, we set $\alpha$ to 1.
The only parameter to be tuned in our experiments is $\lambda$.
We discuss the tuning of this parameter in \S\ref{sec:exp_setup}.
The representation of a synset $y_i$ can be accordingly calculated as the centroid of the vectors of its associated word senses, i.e.,
\begin{equation} 
\{\frac{v_{y_i}}{\lVert v_{y_i} \lVert} : v_{y_i} = \sum_{s \in \mu(y_i)} \hat{v}^*_{s}, \hat{v}^*_{s} = \frac{v^*_{s}}{\lVert v^*_{s} \lVert}\}. \end{equation}

\begin{table}[t!]
\setlength{\tabcolsep}{2pt}
\begin{center}
\small
\scalebox{0.85}
{
\begin{tabular}{ll}
\toprule

\bf \#      &       \bf Closest words           \\
\midrule
\multirow{2}{*}{1}           
            &   crappie, trout, guitar, shad, walleye, bassist, angler, catfish,\\
            &   trombone, percussion, piano, drummer, saxophone, jigs, fish  \\
\midrule
\multirow{2}{*}{2}           
            &   baritone, piano, guitar, trombone, saxophone, cello, percussion, \\
            &   tenor, saxophonist, clarinet, pianist, vocals, solos, harmonica \\
\midrule
\multirow{2}{*}{3}
            &   fish, trout, shrimp, anglers, fishing, bait, guitar, salmon,  \\
            &   shark, fisherman, lakes, seafood, drummer, whale, fisheries     \\


\bottomrule

\end{tabular}
}
\end{center}
\caption{\label{table:sample_output} Ten most similar words to the word \textit{bass} (\#1) and two of its senses: music (\#2) and fish (\#3).}
\end{table}

\begin{figure}
\begin{center}
\includegraphics[scale=1.4]{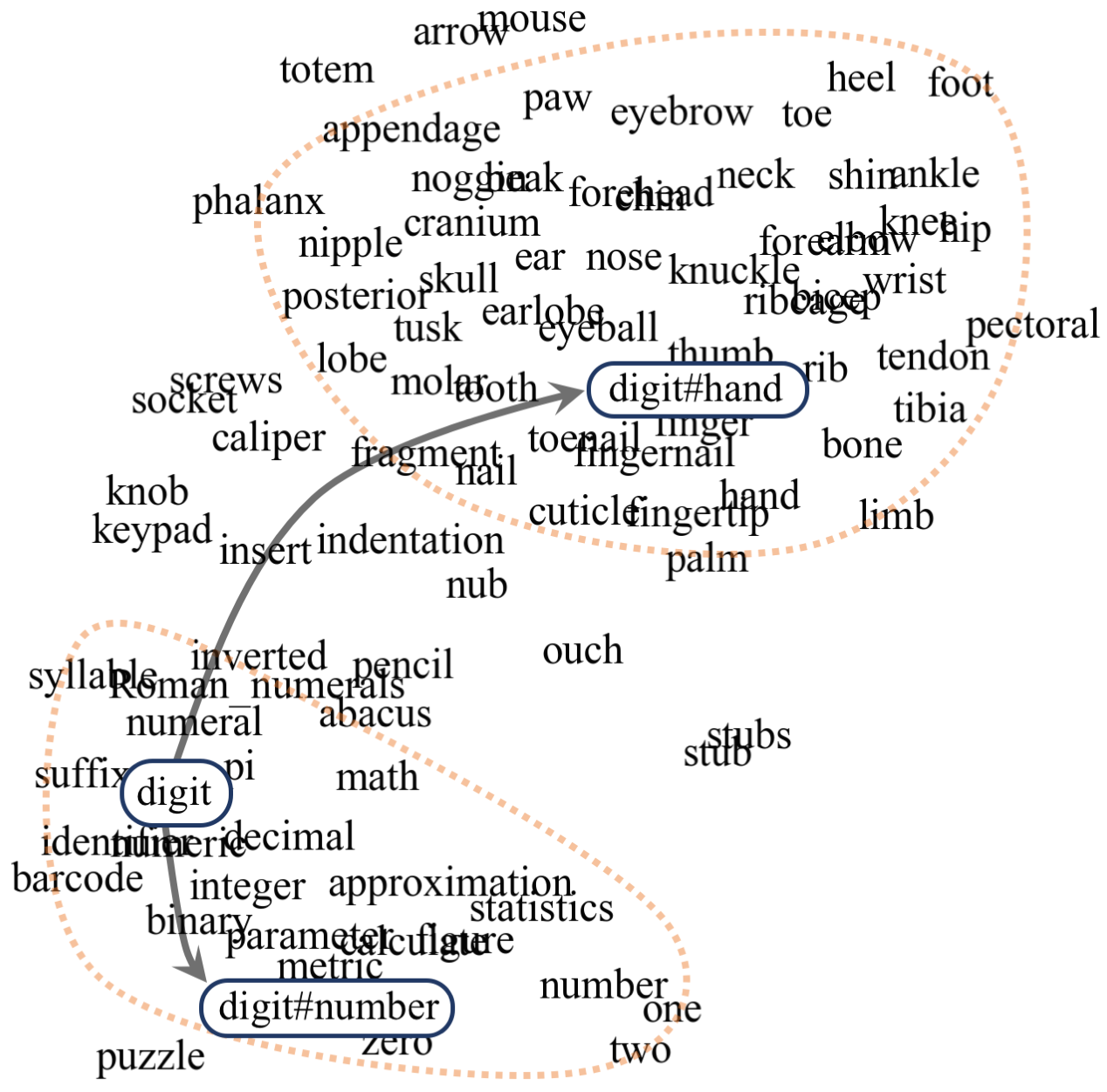}
\end{center}
    \caption{\label{fig:illustration}The illustration of the word \textit{digit} and two of its computed senses in our unified 2-$d$ semantic space.}
\end{figure}

As a result of this procedure, we obtain the set $\mathcal{V}^*$ of $n$ sense representations in the same semantic space as word representations in $\mathcal{V}$.
In fact, we now have a unified semantic space which enables a direct comparison of the two types of linguistic items.
In \S\ref{sec:clss} we evaluate our approach in the word to sense similarity measurement framework.

We show in Table \ref{table:sample_output} the closest words to the word \textit{bass} and two of its senses, music and fish,\footnote{The first and fourth senses in WordNet 3.0, respectively defined as ``the lowest part of the musical range" and ``the lean flesh of a saltwater fish of the family Serranidae."} in our unified semantic space.
We can see in row \#1 a mixture of both meanings when the word representation is used whereas the closest words to the senses (rows \#2 and \#3) are mostly in-domain and specific to the corresponding sense.

To exhibit another interesting property of our sense representation approach, we depict in Figure \ref{fig:illustration} the word \textit{digit} and its numerical and anatomical senses (from the example in Table \ref{table:alg1_output}) in a 2-$d$ semantic space, along with a sample set of words in their proximity.\footnote{We used the t-SNE algorithm \cite{t-SNE} for dimensionality reduction.}
We can see that the word \textit{digit} is placed in the semantic space in the neighbourhood of words from the numerical domain (lower left of the figure), mainly due the dominance \cite{SandersonRijsbergen:1999} of this sense in the general-domain corpus on which the word embeddings in our experiments were trained (cf. \S\ref{sec:pretrained_embed}).
However, upon de-conflation, the emerging anatomical sense of the word is shifted towards the region in the semantic space which is occupied by anatomical words (upper right of the figure).
A clustering-based sense representation technique would have failed in accurately representing the infrequent anatomical meaning of \textit{digit} by analyzing a general domain corpus (such as the one used here).
But our sense representation technique, thanks to its proper usage of knowledge from a sense inventory, is effective in unveiling and accurately modeling less frequent or domain-specific senses of a given word.

Please note that any vector space model representation technique can be used for the pre-training of word representations in $\mathcal{V}$.
Also, the list of sense biasing words can be obtained for larger sense inventories, such as FreeBase \cite{Bollackeretal:2008} or BabelNet \cite{NavigliPonzetto:12aij}.
We leave the exploration of other ways of computing sense biasing words to the future work.

\section{Experiments}

We benchmarked our sense representation approach against several recent techniques on two standard tasks: word similarity (\S\ref{sec:word_similarity}), for which we evaluate on both in-isolation and in-context similarity datasets, and cross-level semantic similarity (\S\ref{sec:clss}).

\subsection{Experimental setup}
\label{sec:exp_setup}

\paragraph{Pre-trained word representations.}
\label{sec:pretrained_embed}
As our word representations, we used the 300-$d$ Word2vec \cite{Mikolovetal:2013} word embeddings trained on the Google News dataset\footnote{\scriptsize{\url{https://code.google.com/archive/p/word2vec/}}} mainly for their popularity across different NLP applications.
However, our approach is equally applicable to any count-based representation technique \cite{BaroniLenci:2010,TurneyPantel:2010} or any other embedding approach \cite{Penningtonetal:2014,Lecun2015deep}.
We leave the evaluation and comparison of various word representation techniques with different training approaches, objectives, and dimensionalities to the future work.

\paragraph{Parameter tuning.}
Recall from \S\ref{sec:learn_rep} that our procedure for learning sense representations needs only one parameter to be tuned, i.e., $\lambda$.
We did not perform an extensive tuning on the value of this parameter and set its value to \nicefrac{1}{5} after trying four different values (1, \nicefrac{1}{2}, \nicefrac{1}{5}, and \nicefrac{1}{10}) on a small validation dataset. 
We leave the more systematic tuning of the parameter and the choice of alternative decay functions (cf. \S\ref{sec:learn_rep}) to the future work.

\paragraph{The size of the sense biasing words lists.}
\label{sec:down_sizing}
Also recall from \S\ref{sec:filtering_words} that the extracted lists of sense biasing words were originally as large as the total number of unique strings in WordNet (around 150K in ver. 3.0).
But, given that we use an exponential decay function in our learning algorithm (cf. \S\ref{sec:learn_rep}), the impact of the low-ranking words in the list is negligible.
In fact, we observed that taking a very small portion of the top-ranking words, i.e., the top 25, produces similarity scores that are on par with those generated when the full lists were considered.
Therefore, we experimented with the down-sized lists which enabled us to generate very quickly sense representations for all word senses in WordNet.

\subsection{Word similarity}
\label{sec:word_similarity}

\paragraph{Comparison systems.}
We compared our results against nine other sense representation techniques: the WordNet-based approaches of \newcite{PilehvarNavigli:2015aij}, \newcite{Chenetal:2014}, \newcite{RotheSchutze:2015}, \newcite{Jauharetal:2015}, and \newcite{Iacobaccietal:2015} and the clustering-based approaches of \newcite{Huangetal:2012}, \newcite{Tianetal:2014}, \newcite{Neelakantanetal:2014}, and \newcite{Liuetal:2015} (please see \S\ref{sec:rel_work} for more details).
We also compared against the approach of \newcite{Faruquietal:2015} which uses knowledge derived from WordNet for improving word representations.
From the different configurations presented in \cite{Faruquietal:2015} we chose the system that uses GloVe \cite{Penningtonetal:2014} with \textit{all} WordNet relations which is their best performing monolingual system.
As for the approach of \newcite{Jauharetal:2015}, we show the results of the EM+RETERO system which performs most consistently across different datasets.

\begin{table*}[t!]
\setlength{\tabcolsep}{20pt}
\renewcommand{\arraystretch}{0.95}
\begin{center}
\small
\scalebox{0.97}
{
\begin{tabular}{llcccc}
\toprule
\multirow{2}{*}{\bf Dataset}   &  \multirow{2}{*}{\bf Approach}  & \multicolumn{2}{c}{\bf Sense-based score} & \multicolumn{2}{c}{\bf Word-based score} \\
&                               &  $r$      &  $\rho$    &  $r$      &   $\rho$   \\
\midrule
\multirow{4}{*}{\rotatebox[origin=c]{0}{MEN-3K}}&

\newcite{Iacobaccietal:2015}    &      $-$      &  \bf 80.5    &        $-$     &   66.5    \\
& \textsc{DeConf}               &  \bf 78.0     &   78.6       &      72.3      &   73.2    \\
& \newcite{Faruquietal:2015}    &      $-$      &   75.9       &        $-$     &   73.7    \\
& \newcite{PilehvarNavigli:2015aij}&   61.7    &   66.6       &   $-$         &   $-$     \\


\midrule
\multirow{4}{*}{\rotatebox[origin=c]{0}{RG-65}} &

\textsc{DeConf}                     &  \bf  90.5    &   \bf  89.6   &   77.2        &   76.1       \\
& \newcite{Iacobaccietal:2015}      &       $-$     &   87.1        &   $-$         &   73.2        \\
& \newcite{Faruquietal:2015}        &       $-$     &   84.2        &   $-$         &   76.7        \\
& \newcite{PilehvarNavigli:2015aij}&       80.2    &   84.3        &   $-$         &   $-$     \\

\midrule
\multirow{3}{*}{\rotatebox[origin=c]{0}{YP-130}} &

 \newcite{PilehvarNavigli:2015aij}& \bf   79.0      &  \bf   71.0     &   $-$         &   $-$     \\
& \textsc{DeConf}                     &    72.9         &     69.5        &   58.0        &   55.9      \\
& \newcite{Iacobaccietal:2015}      &       $-$       &   63.9          &   $-$         &   34.3      \\

\midrule
\multirow{2}{*}{\rotatebox[origin=c]{0}{SimLex-999}} &
 \textsc{DeConf}                   &    54.2      &     51.7        &   45.4        &   44.2    \\
& \newcite{PilehvarNavigli:2015aij}&    43.4      &     43.6        &   $-$         &   $-$     \\

\bottomrule

\end{tabular}
}
\end{center}
\caption{\label{table:results}Pearson ($r \times 100$) and Spearman ($\rho \times 100$) correlation scores on four standard word similarity benchmarks. For each benchmark, we show the results reported by any of the comparison systems along with the scores for their corresponding initial word representations (word-based).}
\end{table*}

\begin{table}[t!]
\renewcommand{\arraystretch}{0.95}
\begin{center}
\small
\scalebox{0.95}
{
\begin{tabular}{lcc}
\toprule
\multirow{2}{*}{\bf Approach}  &  \multicolumn{2}{c}{\bf Score} \\
                                       &   AvgSim      &   AvgSimC     \\
\midrule
\textsc{DeConf}                                  &\bf  70.8      &\bf  71.5 \\
\newcite{RotheSchutze:2015} (best)               &     68.9      &     69.8 \\    
\newcite{Neelakantanetal:2014} (best)            &     67.3      &     69.3 \\
\newcite{Chenetal:2014}                          &     66.2      &     68.9 \\
\newcite{Liuetal:2015} (best)                    &     $-$       &     68.1 \\
\newcite{Huangetal:2012}                         &     62.8      &     65.7 \\
\newcite{Tianetal:2014} (best)                   &     $-$       &     65.7 \\
\newcite{Iacobaccietal:2015}                     &     62.4      &     $-$  \\
\newcite{Jauharetal:2015}                        &      $-$      &     58.7 \\
\cmidrule(lr){2-3}
\textit{Initial word vectors}                    & \multicolumn{2}{c}{65.1}         \\

\bottomrule

\end{tabular}
}
\end{center}
\caption{\label{table:scws_results}Spearman correlation scores ($\rho \times 100$) on the Stanford Contextual Word Similarity (SCWS) dataset. We report the AvgSim and AvgSimC scores (cf. \S\ref{sec:word_similarity}) for each system, where available.}
\end{table}

\paragraph{Benchmarks.}
As our word similarity benchmark, we considered five datasets: RG-65 \cite{RG65:1965}, YP-130 \cite{Yang06verbsimilarity}, MEN-3K \cite{Men3k:2014}, SimLex-999 \cite[SL-999]{Hilletal:2015}, and Stanford Contextual Word Similarity \cite[SCWS]{Huangetal:2012}.
The latter benchmark provides for each word a context that triggers a specific meaning of the word, making it very suitable for the evaluation of sense representation techniques.
For each of the datasets, we list the results that are reported by any of our comparison systems.

\paragraph{Similarity measurement.}
For the SCWS dataset, we follow the past works \cite{ReisingerMooney:2010,Huangetal:2012} and report the results according to two system configurations: 
(1) AvgSim: where the similarity between two words is computed as the average of all the pairwise similarities between their senses, and (2) AvgSimC: where each pairwise sense similarity is weighted by the relevance of each sense to its corresponding context.
For all the other datasets, since words are not provided with any context (they are in isolation), we measure the similarity between two words as that between their most similar senses.
In all the experiments, we use the cosine distance as our similarity measure.

\subsubsection{Experimental results}

Tables \ref{table:scws_results} and \ref{table:results} show the results of our system, \textsc{DeConf}, and the comparison systems on the SCWS and the other four similarity datasets, respectively.
In both tables we also report the word vectors baseline, whenever they are available, which is computed by directly comparing the corresponding word representations of the two words ($\in \mathcal{V}$).
Please note that the word-based baseline does not apply to the approach of \newcite{PilehvarNavigli:2015aij} as it is purely based on the semantic network of WordNet and does not use any pre-trained word embeddings.

We can see from the tables that our sense representations obtain considerable improvements over those of words across the five datasets.
This highlights the fact that the de-conflation of word representations into those of their individual meanings has been highly beneficial.
On the SCWS dataset, \textsc{DeConf} outperforms all the recent state-of-the-art sense representation techniques (in their best settings) which proves the effectiveness of our approach in capturing the semantics of specific meanings of the words.
The improvement is consistent across both system configurations (i.e., AvgSim and AvgSimC).
Moreover, the state-of-the-art WordNet-based approach of \newcite{RotheSchutze:2015} uses the same initial word vectors as \textsc{DeConf} does (cf. \S\ref{sec:pretrained_embed}).
Hence, the improvement we obtain indicates that our approach has made better use of the sense-specific knowledge encoded in WordNet.

As seen in Table \ref{table:results} our approach shows competitve performance on the other four datasets.
The YP-130 dataset focuses on verb similarity, whereas SimLex-999 contains verbs and adjectives and MEN-3K has word pairs with different parts of speech (e.g., a noun compared to a verb).
The results we obtain on these datasets exhibit the reliability of our approach in modeling non-nominal word senses.

\begin{table*}[t!]
\setlength{\tabcolsep}{10.5pt}
\renewcommand{\arraystretch}{0.9}
\begin{center}
\small
{
\begin{tabular}{lcccccccc}
\toprule
\multirow{2}{*}{\bf System}     &   \multicolumn{2}{c}{\bf MaxSim}  & \multicolumn{2}{c}{\bf AvgSim}  &  \multicolumn{2}{c}{\bf S2W} & \multicolumn{2}{c}{\bf S2A}\\
\cmidrule(lr){2-3}
\cmidrule(lr){4-5}
\cmidrule(lr){6-7}
\cmidrule(lr){8-9}
                                  &  \bf     $r$  &  \bf $\rho$   &  \bf     $r$  &  \bf $\rho$   &  \bf     $r$  &  \bf $\rho$   &  \bf     $r$  &  \bf $\rho$ \\
\midrule
\textsc{DeConf}$^*$               &\bf  36.4      &\bf   37.6     &\bf  36.8      &\bf   38.8     &\bf  34.9      &\bf   35.6     &\underline{\bf 37.5}      &\underline{\bf   39.3}  \\
  \newcite{RotheSchutze:2015}$^*$ &     34.0      &      33.8     &     34.1      &      33.6     &     33.4      &      32.0     &\underline{35.4}      &      \underline{34.9}  \\
  \newcite{Iacobaccietal:2015}$^*$&     19.1      &      21.5     &     21.3      &\underline{24.2}     & \underline{22.7}      &      21.7     &     19.5      &      21.1  \\
  \newcite{Chenetal:2014}$^*$     &     17.7      &      18.0     &     17.2      &      16.8     & \underline{27.7}      & \underline{26.7}     &     17.9      &      18.8  \\
\cmidrule(lr){1-9}
 \textsc{DeConf}                  &     35.5      &      36.4     &     36.2      &      38.0     &     34.9      &      35.6     & \underline{36.8}      &      \underline{38.4}  \\
 \newcite{PilehvarNavigli:2015aij}&    19.4      &      23.8     & \underline{21.2}  &     \underline{26.0}      &     $-$       &     $-$       &     $-$       &     $-$       \\
  \newcite{Iacobaccietal:2015}    &     19.0      &      21.5     &     20.9      & \underline{23.2}     &   \underline{22.3}      &      20.6     &     19.2      &      20.4  \\
\bottomrule

\end{tabular}
}
\end{center}
\caption{\label{table:results_clss} Evaluation results on the word to sense similarity test dataset of the SemEval-14 task on Cross-Level Semantic Similarity, according to Pearson ($r \times 100$) and Spearman ($\rho \times 100$) correlations. We show results for four similarity computation strategies (see \S\ref{sec:clss}). The best results per strategy are shown in bold whereas they are underlined for the best strategies per system. Systems marked with $*$ are evaluated on a slightly smaller dataset (474 of the original 500 pairs) so as to have a fair comparison with \protect\newcite{RotheSchutze:2015} and \protect\newcite{Chenetal:2014} that use older versions of WordNet (1.7.1 and 1.7, respectively).}
\end{table*}

\subsubsection{Discussion}
The similarity scale of the SimLex-999 dataset is different from our other word similarity benchmarks in that it assigns relatively low scores to antonymous pairs.
For instance, \textit{sunset-sunrise} and \textit{man-woman} in this dataset are assigned the respective similarities of 2.47 and 3.33 (in a $[0,10]$ similarity scale) which is in the same range as the similarity between word pairs with slight domain relatedness, such as \textit{head-nail} (2.47), \textit{air-molecule} (3.05), or \textit{succeed-try} (3.98).
In fact we observed that tweaking the similarity scale of our system in a way that it diminishes the similarity scores between antonyms can result in significant performance improvement on this dataset.
To this end, we performed an experiment in which the similarity of a word pair was simply divided by five whenever the two words belonged to synsets that were linked by the antonymy relation. We observed that the performance on the SimLex-999 dataset increased to 61.1 (from 54.2) and 59.0 (from 51.7) according to Pearson ($r \times 100$) and Spearman ($\rho \times 100$) correlation scores, respectively.

\subsection{Cross-Level semantic similarity}
\label{sec:clss}

In addition to the word similarity benchmark, we evaluated the performance of our representations in the cross-level semantic similarity measurement framework.
To this end, we opted for the SemEval-2014 task on Cross-Level Semantic Similarity \cite[CLSS]{Jurgensetal:2014}.
The word to sense similarity subtask of this task, with 500 instances in its test set, provides a suitable benchmark for the evaluation of sense representation techniques.

For a word sense $s$ and a word $w$, we compute the similarity score according to four different strategies:
the similarity of $s$ to the most similar sense of $w$ (\textbf{MaxSim}), the average similarity of $s$ to individual senses of $w$ (\textbf{AvgSim}), the direct similarity of $s$ to $w$ when the latter is modeled as its word representation (Sense-to-Word or \textbf{S2W}) or as the centroid of its senses' representations (Sense to aggregated word senses or \textbf{S2A}).
For this task, we can only compare against the publicly-available sense representations of \newcite{Iacobaccietal:2015}, \newcite{RotheSchutze:2015}, \newcite{PilehvarNavigli:2015aij} and \newcite{Chenetal:2014} which are linked to the WordNet sense inventory.

\subsubsection{Experimental results}
\label{sec:clss_results}

Table \ref{table:results_clss} shows the results on the word to sense dataset of the SemEval-2014 CLSS task, according to Pearson ($r$) and Spearman ($\rho$) correlations and for the four strategies.
As can be seen from the low overall performances, the task is a very challenging benchmark with many WordNet out-of-vocabulary or slang terms and rare usages.
Despite this, \textsc{DeConf} provides consistent improvement over the comparison sense representation techniques according to both measures and for all the strategies.

Across the four strategies, S2A proves to be the most effective for \textsc{DeConf} and the representations of \newcite{RotheSchutze:2015}.
The representations of \newcite{Chenetal:2014} perform best with the S2W strategy whereas those of \newcite{Iacobaccietal:2015} do not show a consistent trend with relatively low performance across the four strategies.
Also, a comparison of our results across the S2W and S2A strategies reveals that a word's aggregated representation, i.e., the centroid of the representations of its senses, is more accurate than its original word representation.

Our analysis showed that the performances of the approaches of \newcite{RotheSchutze:2015} and \newcite{Iacobaccietal:2015} were hampered partly due to their limited coverage.
In fact, the former was unable to model around 35\% of the synsets in WordNet 1.7.1, mainly for its shallow exploitation of knowledge from WordNet, whereas the latter approach did not cover around 15\% of synsets in WordNet 3.0.
\newcite{Chenetal:2014} provide near-full coverage for word senses in WordNet. However, the relatively low performance of their system shows that the usage of glosses in WordNet and the automated disambiguation have not resulted in accurate sense representations.
Thanks to its deep exploitation of the underlying resource, our approach provides full coverage over all word senses and synsets in WordNet.

The three best-performing systems in the task are Meerkat\_Mafia \cite{MeerkatMafia:2014} ($r=37.5$, $\rho=39.3$), SimCompass \cite{SimCompass:2014} ($r=35.4$, $\rho=34.9$), and SemantiKLUE \cite{SemantiKLUE:2014} ($r=17.9$, $\rho=18.8$).
Please note that these systems are specifically designed for the cross-level similarity measurement task.
For instance, the best-ranking system in the task leverages a compilation of several dictionaries, including The American Heritage Dictionary, Wiktionary and WordNet, in order to handle slang terms and rare usages, which leads to its competitive performance \cite{MeerkatMafia:2014}.


\section{Related Work}
\label{sec:rel_work}

Learning semantic representations for individual senses of words has been an active area of research for the past few years.
Based on the way they view the problem, the recent techniques can be classified into two main branches: 
(1) those that, similarly to our work, extract knowledge from external sense inventories for learning sense representations; and 
(2) those techniques that cluster the contexts in which a word appears in a given text corpus and learn distinct representations for individual clusters. 

Examples for the first branch include the approaches of  \newcite{Chenetal:2014}, \newcite{Jauharetal:2015} and \newcite{RotheSchutze:2015}, all of which use WordNet as an external resource and obtain sense representations for this sense inventory. 
\newcite{Chenetal:2014} uses the content words in the definition of a word sense and WSD.
However, the sole usage of glosses as sense-distinguishing contexts and the non-optimal WSD make the approach inaccurate, particularly for highly polysemous words with similar senses and for word senses with short definitions.
Similarly, \newcite{RotheSchutze:2015} use only polysemy and synonymy properties of words in WordNet along with a small set of semantic relations. 
This significantly hampers the reliability of the technique in providing high coverage (discussed further in \S\ref{sec:clss_results}). 
Our approach improves over these works by exploiting deep knowledge from the semantic network of WordNet, coupled with an effective training approach.
ADW \cite{PilehvarNavigli:2015aij} is another WordNet-based approach which exploits only the semantic network of this resource an obtains interpretable sense representations.
Other work in this branch include SensEmbed \cite{Iacobaccietal:2015} and Nasari \cite{camachocolladosetal:2015,CamachoPilehvarNavigli:2016aij} which are based on the BabelNet sense inventory \cite{NavigliPonzetto:12aij}.
The former technique first disambiguates words in a given corpus with the help of a knowledge-based WSD system and then uses the generated sense-annotated corpus as training data for Word2vec.
Nasari combines structural knowledge from the semantic network of BabelNet with corpus statistics derived from Wikipedia for representing BabelNet synsets.
However, the approach falls short of modeling non-nominal senses as Wikipedia, due to its very encyclopedic nature, does not cover verbs, adjectives, or adverbs.

The second branch, which is usually referred to as \textit{multi-prototype} representation, is often associated with clustering.
\newcite{ReisingerMooney:2010} proposed one of the recent pioneering techniques in this branch.
Other prominent work in the category include topical word embeddings \cite{Liuetal:2015} which use latent topic models for assigning topics to each word in a corpus and learn topic-specific word representations, and the technique proposed by \newcite{Huangetal:2012} which incorporates ``global document context."
\newcite{Tianetal:2014} modified the Skip-gram model in order to learn multiple embeddings for each word type.
Despite the fact that these techniques do not usually take advantage of the knowledge encoded in structured knowledge resource, they generally suffer from two disadvantages. 
The first limitation is that they usually make an assumption that a given word has a fixed number of senses, ignoring the fact that polysemy is highly dynamic across words that can range from monosemous to highly ambiguous with dozens of associated meanings \cite{McCarthyetal:2016}.
\newcite{Neelakantanetal:2014} tackled this issue by estimating the number of senses for a word type during the learning process.
However, all techniques in the second branch suffer from another disadvantage that their computed sense representations are not linked to any sense inventory, a linking which itself would require the existence of high coverage sense-annotated data \cite{Agirreetal:2006}. 

Another notable line of research incorporates knowledge from external resources, such as PPDB \cite{PPDB} and WordNet, to improve word embeddings \cite{YuDredze:2014,Faruquietal:2015}. 
Neither of the two techniques however provide representations for word senses.

\section{Conclusions}
We put forward a sense representation technique, namely \textsc{DeConf}, that provides multiple advantages in comparison to the recent state of the art: 
(1) the number of word senses in our technique is flexible and the computed representations are linked to word senses in WordNet; 
(2) \textsc{DeConf} is effective in providing accurate representation of word senses, even for those senses that do not usually appear frequently in generic text corpora;
and (3) our approach is general in that it can be readily applied to any set of word representations and any semantic network without the need for extensive parameter tuning.
Our experimental results showed that \textsc{DeConf} can outperform recent state of the art on several datasets across two tasks.
We release our computed representations for around 118K synsets and 205K word senses in WordNet 3.0 at \url{https://github.com/pilehvar/deconf}.
As future work, we plan to investigate the possibility of using larger semantic networks, such as FreeBase and BabelNet, which would also allow us to apply the technique to languages other than English.
We also plan to evaluate the performance of our approach with other decay functions as well as with other initial word representations.

\section*{Acknowledgments}
The authors gratefully acknowledge the support of the MRC grant No. MR/M025160/1 for PheneBank.

\bibliography{emnlp2016}
\bibliographystyle{emnlp2016}

\end{document}